\documentclass[sigconf]{acmart}
\AtBeginDocument{%
  }

\setcopyright{acmlicensed}
\copyrightyear{2018}
\acmYear{2018}
\acmDOI{XXXXXXX.XXXXXXX}
\acmConference[Conference acronym 'XX]{Make sure to enter the correct
  conference title from your rights confirmation email}{June 03--05,
  2018}{Woodstock, NY}
\acmISBN{978-1-4503-XXXX-X/2018/06}

\usepackage{enumitem}

\begin{document}

\title{Participatory Evolution of Artificial Life Systems via Semantic Feedback
}


\author{Shuowen Li}
\authornote{Both authors contributed equally to this research.}
\email{lsw23@mails.tsinghua.edu.cn}
\affiliation{%
  \institution{Tsinghua University}
  \city{Beijing}
  \country{China}
}

\author{Kexin Wang}
\authornotemark[1]
\authornote{Project Lead.}
\email{kexinw@mit.edu}
\orcid{0009-0001-2835-4148}
\affiliation{%
  \institution{MIT CSAIL}
  \city{Cambridge}
  \state{MA}
  \country{USA}
}

\author{Minglu Fang}
\email{2221305075@caa.edu.cn}
\affiliation{%
  \institution{China Academy of Art}
  \city{Hangzhou}
  \state{Zhejiang}
  \country{China}
}

\author{Danqi Huang}
\email{hdq24@mails.tsinghua.edu.cn}
\affiliation{%
  \institution{Tsinghua University}
  \city{Beijing}
  \country{China}
}

\author{Ali Asadipour}
\email{ali.asadipour@rca.ac.uk}
\orcid{0000-0003-0159-3090}
\affiliation{%
  \institution{Royal College of Art, Computer Science Research Centre}
  \city{London}
  \country{United Kingdom}
}

\author{Haipeng Mi}
\email{mhp@tsinghua.edu.cn}
\authornotemark[3]
\affiliation{%
  \institution{Tsinghua University}
  \city{Beijing}
  \country{China}
}

\author{Yitong Sun}
\email{yitong.sun@network.rca.ac.uk}
\orcid{0000-0002-9469-7157}
\authornote{Corresponding author.}
\affiliation{%
  \institution{Royal College of Art, Computer Science Research Centre}
  \city{London}
  \country{United Kingdom}
}

\renewcommand{\shortauthors}{Shuowen Li, Kexin Wang et al.}

\begin{abstract}

We present a semantic feedback framework that enables natural language to guide the evolution of artificial life systems. Integrating a prompt-to-parameter encoder, a CMA-ES optimizer, and CLIP-based evaluation, the system allows user intent to modulate both visual outcomes and underlying behavioral rules. Implemented in an interactive ecosystem simulation, the framework supports prompt refinement, multi-agent interaction, and emergent rule synthesis. User studies show improved semantic alignment over manual tuning and demonstrate the system’s potential as a platform for participatory generative design and open-ended evolution.

\end{abstract}


\begin{teaserfigure}
\begin{center}
  \includegraphics[width=0.8\textwidth]{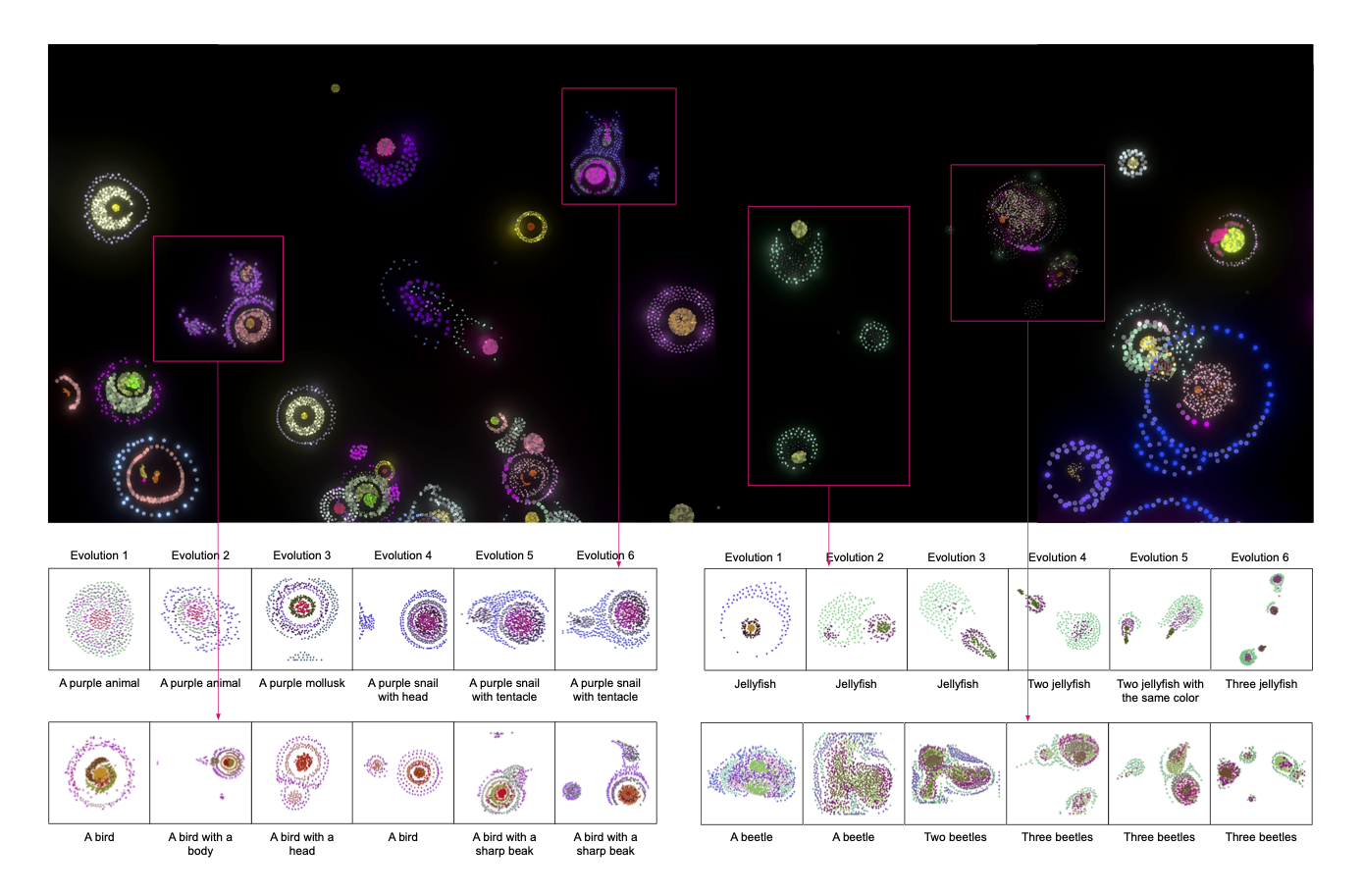}
  \caption{Evolutionary Pathways of Digital Lifeforms and Schematic Illustration of Ecological States in a Digital Life System.}
  \Description{Enjoying the baseball game from the third-base
  seats. Ichiro Suzuki preparing to bat.}
  \label{fig:teaser}
\end{center}
\end{teaserfigure}

\received{20 February 2007}
\received[revised]{12 March 2009}
\received[accepted]{5 June 2009}

\maketitle

\section{Introduction}
Human interaction with complex systems, including ecological environments, social dynamics, and artificial simulations, presents ongoing challenges for both modeling and creative intervention. While agent-based models and artificial life systems provide formal tools for simulating emergent behavior, they often rely on fixed rule structures and limited input modalities. This makes it difficult to embed high-level conceptual intent into evolving system dynamics.

Recent advances in multimodal AI, particularly vision-language models such as CLIP, have expanded the potential of using natural language to guide generative systems. However, most implementations adopt a static prompt-to-output paradigm in which language acts as a trigger rather than a continuous modulator. This restricts semantic iteration and reduces a system’s ability to adapt behavior in response to evolving user input.

We introduce a semantic feedback-driven framework for evolving artificial life systems, where natural language serves as both expressive input and regulatory signal. Grounded in the Swarm model, our architecture integrates a BERT-based prompt-to-parameter encoder, a CMA-ES evolutionary optimizer, and a CLIP-based semantic evaluation module. This closed-loop pipeline allows language to influence both the visual output and the behavioral rules of the simulation, enabling interpretable, adaptive, and semantically aligned evolution.

To demonstrate the framework’s capabilities, we implemented it in an interactive platform that enables users to co-construct a semantic ecosystem. Participants generate and evolve synthetic agents using language prompts. These agents coexist and interact in a shared artificial environment, exhibiting behaviors shaped by both local rules and emergent collective dynamics. As prompts accumulate, the system analyzes linguistic histories and behavioral trajectories to derive meta-rules that recursively influence subsequent evolution.

We evaluated the system in two settings: a controlled study assessing its alignment with user intent, and a public deployment supporting open-ended co-creation. Results suggest that semantic input enhances controllability, interpretability, and user engagement, positioning the framework as a novel platform for participatory artificial life and expressive complex system design.

\section{Related Work}

\subsection{Consciousness and the Emergence of Artificial Life}

Contemporary theories of consciousness, including relational, ecological, and generative perspectives, challenge traditional assumptions that define consciousness as uniquely human and a priori. Frameworks such as animism \cite{harvey2005animism}, the extended mind \cite{clark1998extended}, and embodied cognition \cite{varela1991embodied} reconceptualize consciousness as a regulatory process emerging from system–environment coupling. Within this view, consciousness operates as a feedback mechanism that enables self-organization, boundary modulation, and adaptive response to perturbations. These principles have inspired the design of artificial systems that embed similar dynamics of self-regulation and emergent behavior.

Agent-based models, particularly those within the artificial life (ALife) tradition, such as Reynolds' Boids \cite{reynolds1987flocks}, cellular automata, and swarm-based ecologies \cite{Sayama2009SwarmChemistry}, offer formal tools for studying emergent complexity through decentralized local interaction. These systems exemplify how simple rules at the individual level can give rise to lifelike collective phenomena. While ALife models have been widely applied in generative art, ecological simulation, and swarm robotics, most rely on predefined rules and do not incorporate semantically adaptive input, limiting their responsiveness to human conceptual intent \cite{park2023generative, yitong}.

This opens new possibilities for language-integrated artificial life systems. Rather than treating language as a symbolic descriptor of output, we explore its function as a continuous regulatory input that is functionally analogous to consciousness. In this role, language actively shapes system evolution in real time. Our work positions semantic feedback as a mediator between user intent and behavioral adaptation, extending the scope of ALife through participatory, linguistically guided evolution.

\subsection{Language Models as Evolutionary Drivers}

Recent advances in multimodal embedding models, particularly CLIP, have greatly improved systems’ capacity to interpret and respond to natural language within generative contexts \cite{tian2022modern}. Platforms such as Midjourney, DALL·E, and Stable Diffusion translate textual prompts into rich visual outputs, while systems like DreamFusion \cite{poole2022dreamfusion} and Text2Mesh \cite{michel2022text2mesh} extend these capabilities into the realm of 3D generation. Despite their sophistication, these platforms follow a predominantly static prompt‑to‑output paradigm, in which user input does not influence the underlying generative mechanisms over time.

At the same time, foundational research in artificial life has begun to integrate language-aligned foundation models. Kumar et al.’s Automated Search for Artificial Life (ASAL) framework leverages vision-language models to explore diverse lifelike behaviors across multiple ALife substrates, such as Boids, Lenia, and cellular automata, discovering open-ended novelty and human-aligned emergence \cite{kumar2024automating}. This work exemplifies how semantic evaluation tools can direct emergent simulations toward biologically evocative patterns, underscoring the potential of language-grounded open-ended search within ALife.

Building on this trajectory, our system enables semantic input to govern not only the visual output but also the behavioral rules in a swarm-based artificial life simulation. Natural language prompts are encoded into parameter vectors that act as both seeds and directional priors. A CLIP-based fitness function quantifies alignment between simulation behaviors and user intent, guiding the CMA-ES optimizer to adapt system dynamics progressively. This closed-loop process realizes a form of language-curated evolution, dynamically evolving agent-based systems in response to continuous semantic signals.

\begin{figure*}[htbp]
  \centering
  \includegraphics[width=\textwidth]{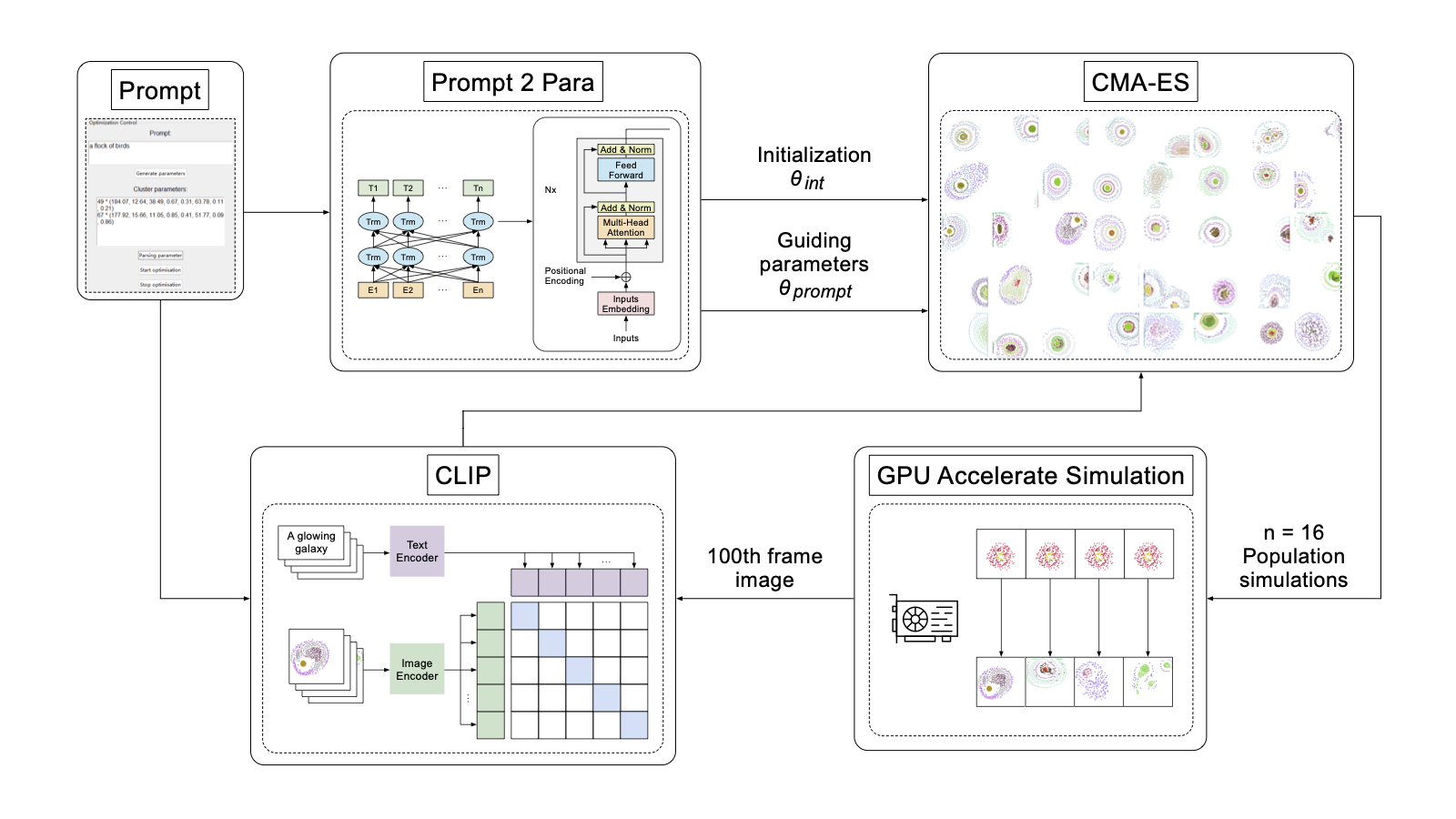}
  \caption{
    System architecture of the semantic feedback-driven evolution framework. Users input a natural language prompt, which is encoded via a text encoder and transformed into guiding parameters for a population-based simulation. A CMA-ES optimizer evaluates each population using CLIP-based semantic similarity between the prompt and simulation outputs. The optimization iteratively updates the parameters to align system behavior with the conceptual intent. The simulation loop is GPU-accelerated, enabling real-time visual feedback over successive generations. The transformer-based embedding model aligns image and text embeddings in a shared semantic space to compute fitness values.
  }
  \label{fig:semantic_loop}
\end{figure*}

\subsection{Participatory Artificial Ecosystems and Generative Art}

Artificial life has long intersected with generative art through systems that embed user input into emergent behavioral frameworks. Pioneering works such as Sommerer and Mignonneau’s Interactive Plant Growing (1994) and A-Volve (1995) \cite{sommerer1997interacting,sommerer2002volve}enabled participants to physically engage with biological or digital organisms, prompting growth and transformation in evolving life forms through touch and gesture-based interaction. Other projects have further explored human input as a driver of artificial evolution. Karl Sims’s Galápagos (1997) \cite{cardenas2014darwin}invited gallery visitors to guide the evolution of virtual creatures by selecting preferred individuals on a screen, using aesthetic preference as a proxy for fitness. In TechnoSphere (1995–2002), users designed digital creatures by specifying morphology, movement style, and behavioral traits through a web-based interface \cite{prophet2001technosphere}. These user-defined organisms were then released into a shared virtual environment where they reproduced, competed for resources, and underwent ecosystem-level selection over time.

While these systems demonstrate how user input can influence artificial life evolution, the modes of interaction are typically constrained to gestures, parameter selections, or predefined symbolic controls. Building on this lineage, our framework introduces natural language as a direct and expressive means of articulating user intent. Compared to fixed input modalities, language is inherently open-ended and semantically rich, making it especially suited for open-ended creative exploration. We propose a closed-loop architecture in which linguistic input continuously modulates the evolution of a particle-based artificial life simulation. The system evolves not only through structural emergence driven by agent interaction but also through ongoing semantic alignment between user prompts and generative behavior. We implement this framework in a publicly accessible interactive platform, demonstrating its potential as a participatory system for language-driven artificial ecosystems and generative art.

\begin{figure*}[htbp]
  \centering
  \includegraphics[width=\textwidth]{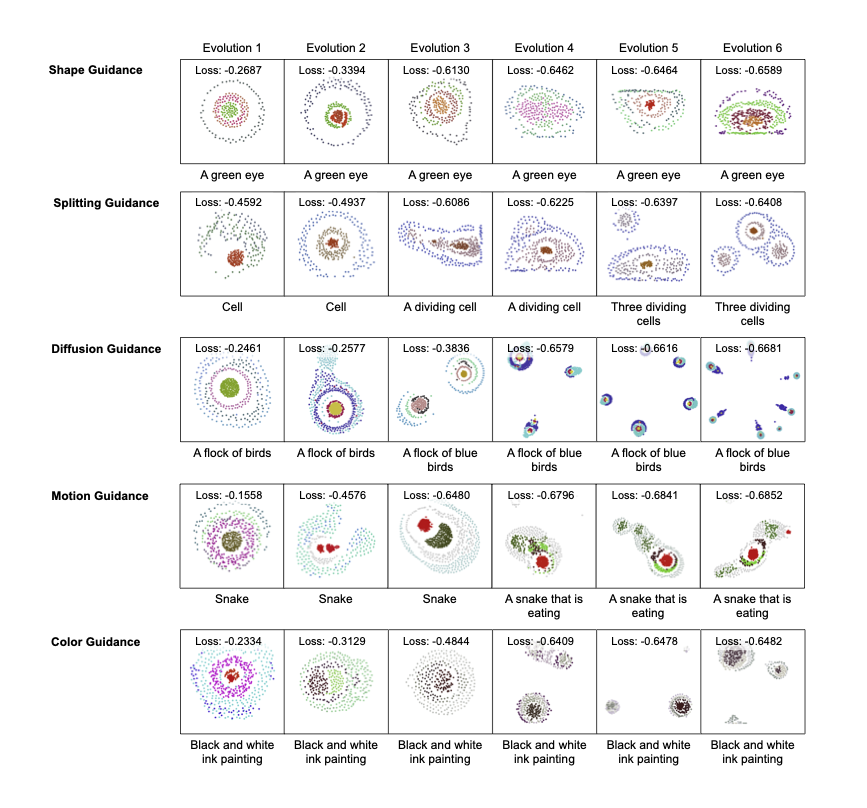}
  \caption{
    Evolutionary trajectories under different semantic prompts. Each row shows a generative sequence across six evolutionary iterations based on a distinct natural language prompt. From top to bottom: "Black and white ink painting," "A snake that is eating," "Three dividing cells," "A green eye," and "A flock of blue birds." The loss values under each image reflect semantic distance between the generated output and the intended concept, as computed via CLIP-based embedding comparison. Prompts guide system behavior through a closed-loop optimization process, illustrating how linguistic input iteratively shapes emergent visual patterns.
  }
  \label{fig:prompt_evolution}
\end{figure*}

\begin{figure*}[htbp]
  \centering
  \includegraphics[width=\textwidth]{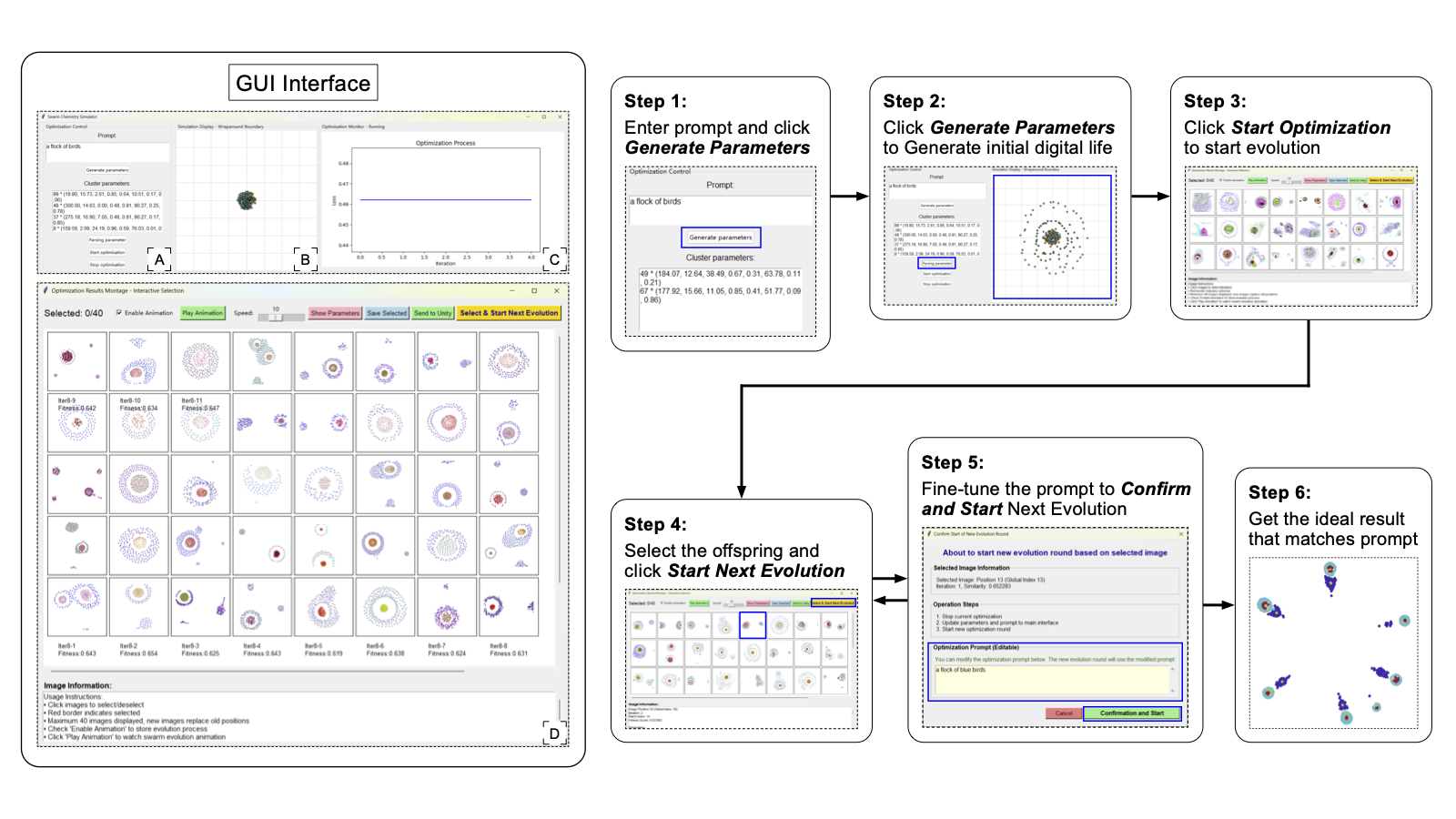}
  \caption{
User interaction workflow within our semantic-driven generative system. (A) Users input a prompt and generate the initial parameter configuration. (B) The system produces an initial simulation based on this prompt. (C) Evolution is initiated through a semantic optimization loop, and users can select preferred outcomes to guide the next generation. (D) Users may refine the prompt at any stage to steer the system toward their conceptual intent. The process continues until the emergent behavior aligns with the user’s evolving goals.
  }
  \label{fig:gui_interface}
\end{figure*}

\section{Methodology}

Unlike image generation tasks, which typically operate under well-defined objectives and constrained solution spaces, the optimization of complex systems presents fundamental challenges, including high-dimensional coupling, nonlinearity, and emergent dynamics. Agent-based simulations such as Swarm exemplify these properties, as they feature intricate dependencies among agents, local interaction rules, and environmental variables, without any explicit or globally optimal state.

Although evolutionary algorithms such as CMA-ES and genetic algorithms have been extensively used in these black-box contexts, they rely heavily on stochastic sampling and are often susceptible to local optima, especially in high-dimensional search spaces \cite{uchida2024covariance}. Recent advances in reinforcement learning have introduced dynamic control strategies \cite{gupta2020networked}, yet few methods integrate real-time semantic feedback or support conceptual intent as a guiding force for system evolution in a flexible and interpretable manner.

To address this gap, we propose a framework for semantically guided optimization of artificial life systems via natural language input. Grounded in the Swarm model, our approach establishes a closed-loop pipeline that links semantic prompts to behavioral rule search and iterative evolutionary refinement. This framework replaces undirected exploration with quasi-random search shaped by linguistic guidance \cite{kumar2024automating}, leveraging the expressive capacity of natural language to introduce meaningful directionality without relying on differentiable objectives.

By treating semantic intent as an active regulatory mechanism within an agent-based environment, our approach facilitates intuitive, high-level intervention in complex system behavior. This reduces the technical barrier to engaging with artificial life simulations and opens new possibilities for collaborative human–machine experimentation, particularly in generative design, interactive art, and interdisciplinary research.

\begin{figure*}[htbp]
  \centering
  \includegraphics[width=\textwidth]{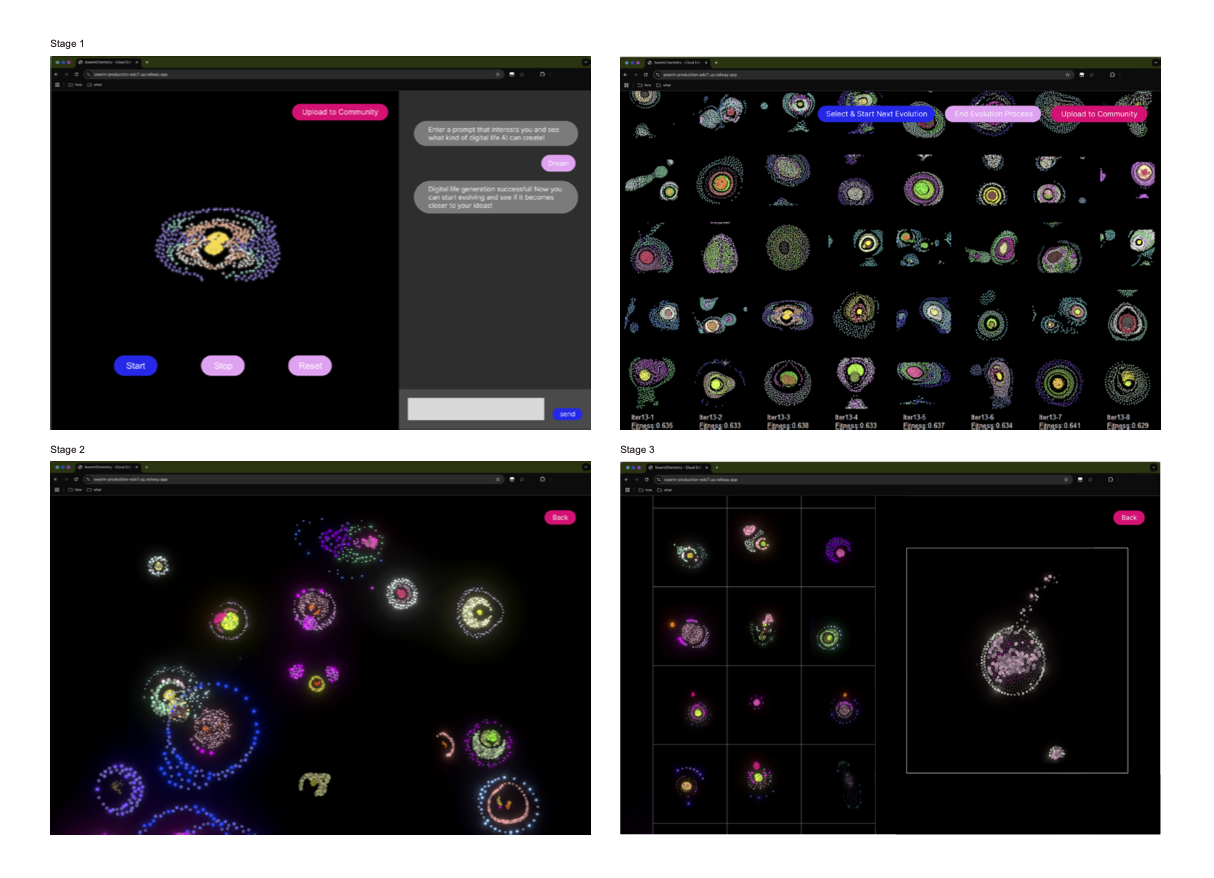}
  \caption{
    System architecture of the semantic feedback-driven evolution framework. Users input a natural language prompt, which is encoded via a text encoder and transformed into guiding parameters for a population-based simulation. A CMA-ES optimizer evaluates each population using CLIP-based semantic similarity between the prompt and simulation outputs. The optimization iteratively updates the parameters to align system behavior with the conceptual intent. The simulation loop is GPU-accelerated, enabling real-time visual feedback over successive generations. The transformer-based embedding model aligns image and text embeddings in a shared semantic space to compute fitness values.
  }
  \label{fig:game_ui}
\end{figure*}

\subsection{Closed-Loop Semantic Feedback Architecture}

As shown in Figure~\ref{fig:semantic_loop}, the system comprises three core components:

\begin{itemize}[topsep=2pt,itemsep=2pt,parsep=0pt]
  \item \textbf{Language Encoding (Prompt2Param):} Natural language prompts are translated into parameter vectors that encode semantic intent.
  \item \textbf{Parameter Optimization (CMA-ES):} Based on these vectors, the system performs evolutionary search to optimize agent-based behavioral configurations.
  \item \textbf{Semantic Evaluation (CLIP):} A vision-language model quantifies the semantic similarity between simulation outputs and the original prompt, supplying feedback to guide the optimization process.
\end{itemize}

Natural language allows users to articulate abstract, affective, or metaphorical goals, such as “expand like a nebula” or “gather like magnets,” which would be difficult to express through numerical inputs alone. To bridge the gap between such high-level expressions and low-level system configurations, we constructed a dataset of 100 natural language prompts paired with corresponding Swarm model parameters. Based on this corpus, we trained a lightweight BERT-based encoder, Prompt2Param, that produces two outputs: an initial configuration vector $\theta_{\text{init}}$ used to initialize the simulation, and a semantic bias vector $\theta_{\text{prompt}}$ that serves as a directional prior throughout the optimization process. This dual-vector design enables both stable initialization and sustained semantic guidance across generations.

For the search process, we employ Covariance Matrix Adaptation Evolution Strategy (CMA-ES), which is particularly well suited for high-dimensional, non-convex, and non-differentiable optimization tasks. Unlike gradient-based methods, CMA-ES operates in black-box settings and does not require access to an explicit loss function. Its flexible, modular structure allows us to incorporate semantic priors derived from language without compromising the algorithm’s stability or convergence properties.

To evaluate how well simulation outcomes align with the user’s semantic intent, we implement a cross-modal fitness function based on OpenAI’s CLIP model. Each simulation is rendered into a sequence of frames, from which representative images are selected and encoded into 512-dimensional visual embeddings using CLIP’s ViT-B/32 encoder. These embeddings are then compared with the prompt’s text embedding using cosine similarity, yielding a semantic fitness score. This score replaces predefined visual metrics with a concept-driven evaluation signal, allowing the system to evolve behaviors that align with human-level descriptions rather than rigid target templates.

\begin{figure*}[htbp]
  \centering
  \includegraphics[width=\textwidth]{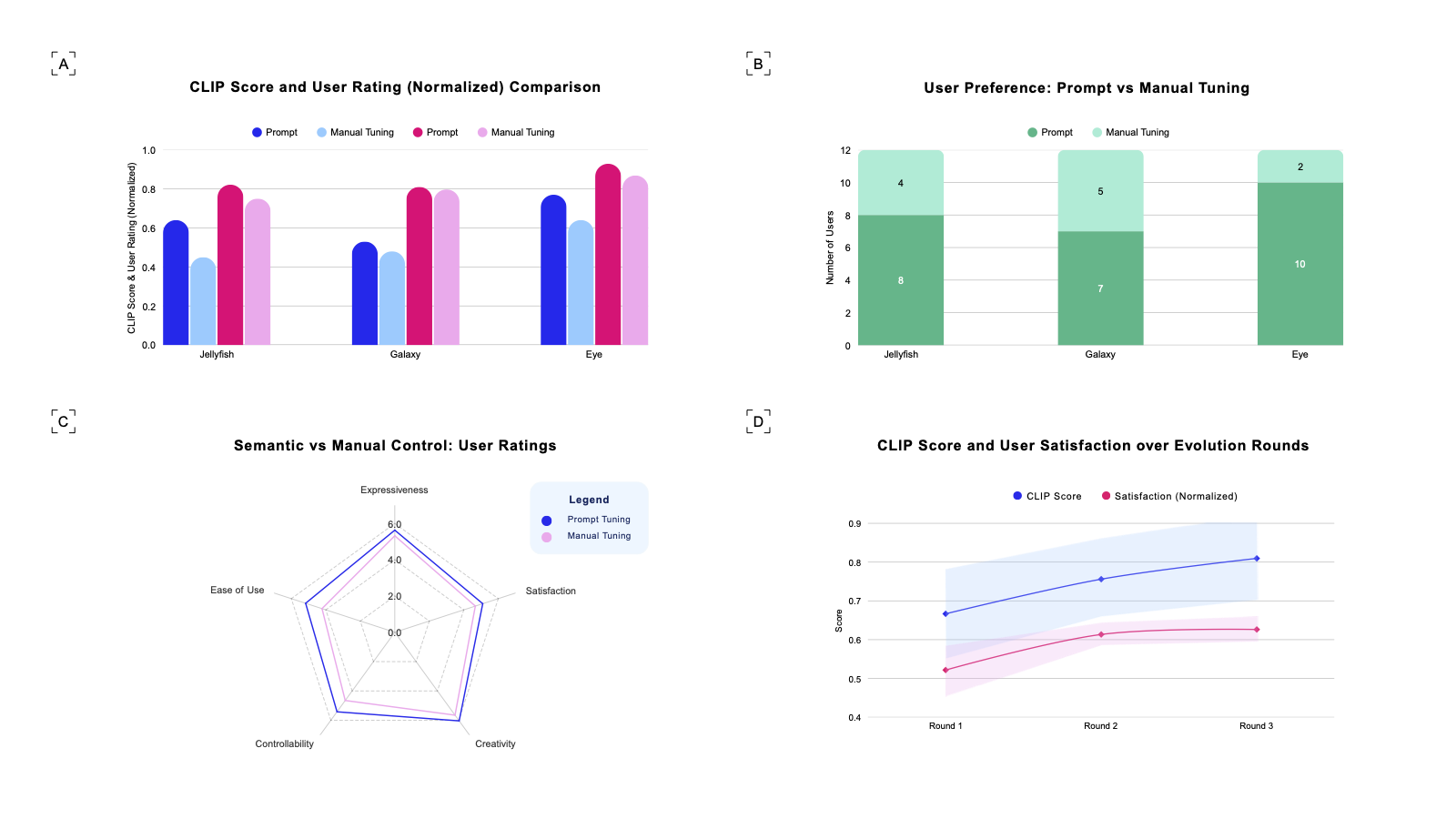}
  \caption{
    Comparative user evaluation of prompt-based semantic tuning versus manual parameter tuning. (A) CLIP score and normalized user ratings (0–1 scale) across three prompts ("Jellyfish," "Galaxy," "Eye") show consistently higher alignment and satisfaction under prompt-based tuning. (B) User preference counts favor prompt tuning over manual tuning for each prompt. (C) Aggregate user ratings across five dimensions—expressiveness, satisfaction, creativity, controllability, and ease of use—demonstrate a consistent advantage for semantic control. (D) Multi-round analysis of prompt refinement indicates progressive improvement in both CLIP score and satisfaction over three iterations.
  }
  \label{fig:user_study}
\end{figure*}

\subsection{Semantic-Driven Evolutionary Workflow}

When a user submits a natural language prompt, the system initiates a semantically guided optimization process. The Prompt2Param module encodes the prompt into two output vectors, the vector $\theta_{\text{init}}$ defines the initial behavioral configuration of the Swarm simulation, including parameters such as speed, neighborhood radius, and cohesion strength. Meanwhile, $\theta_{\text{prompt}}$ functions as a directional prior throughout the evolutionary process, guiding the system toward behaviors aligned with the user’s intended semantics.

Starting from $\theta_{\text{init}}$, the CMA-ES algorithm samples a population of $n = 16$ candidate parameter sets per generation. Each candidate defines a complete behavioral rule set for the Swarm model, including continuous variables such as neighborhood radius, maximum velocity, alignment coefficient, cohesion coefficient, separation coefficient, and noise intensity.

Each parameter set is used to run a full GPU-accelerated simulation. The resulting particle trajectories are rendered into images, which are then processed through the CLIP ViT-B/32 model to generate 512-dimensional visual embeddings. These embeddings are compared against the text embedding of the original prompt using cosine similarity, yielding a semantic fitness score for each candidate.

CMA-ES aggregates the fitness scores across the population to update its sampling distribution, progressively converging toward regions of the parameter space that produce behaviors more semantically aligned with the user’s intent. Throughout this process, $\theta_{\text{prompt}}$ continues to act as a guiding vector, biasing the search in a conceptually coherent direction.

To prevent premature convergence and ensure creative variability, the system introduces mild stochastic noise and parameter mutations at each generation. It also monitors the diversity of visual embeddings across generations. This mechanism preserves visual richness and behavioral diversity, while maintaining semantic consistency with the user-defined prompt. Figure~\ref{fig:prompt_evolution} presents examples of prompt-guided evolution, visually demonstrating how the system refines agent behavior across generations under different semantic conditions.

\subsection{System GUI Integration}

To enhance accessibility and ease of use, we developed a graphical user interface (GUI) designed for artists, designers, and other users without a programming background. Built with Python and Tkinter, the tool allows intuitive interaction with the semantic evolution of complex systems.

Figure~\ref{fig:gui_interface} illustrates the interface layout. Users begin by entering a natural language prompt into the interface. With a single click, the system initiates the full optimization process: the prompt is encoded into simulation parameters, the CMA-ES algorithm performs evolutionary search, and the CLIP model evaluates the semantic alignment of each generation. The interface presents real-time visualization of the Swarm simulation results, enabling users to monitor evolutionary progress, select preferred outcomes, or launch new rounds of optimization based on chosen individuals.

\begin{figure*}[htbp]
  \centering
  \includegraphics[width=\textwidth]{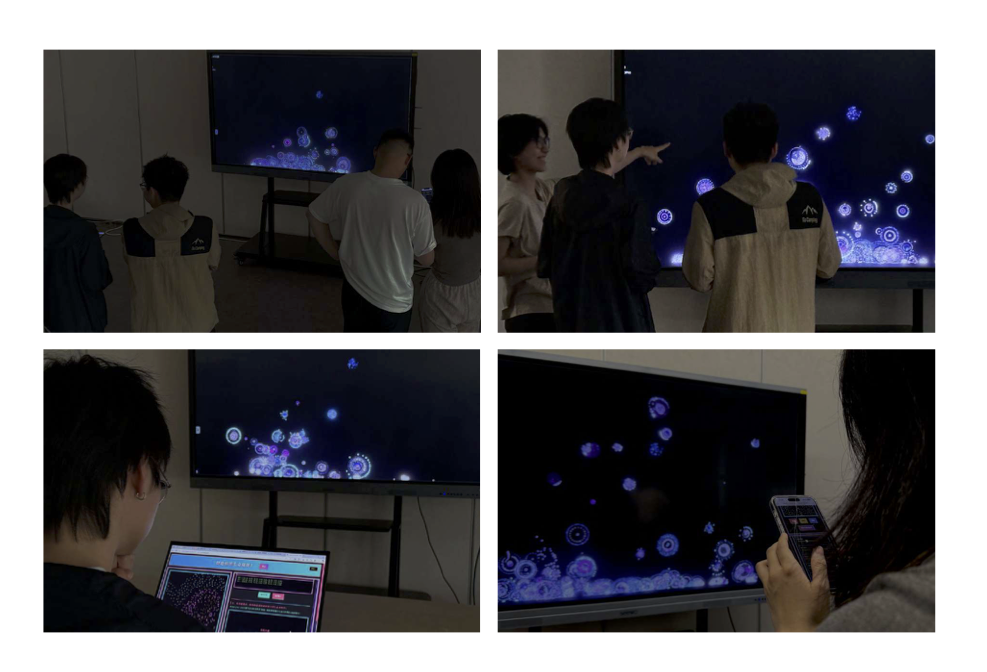}
  \caption{
    On-site interaction with the participatory artificial life game.Participants collaboratively engage with a semantic-driven ecosystem game by submitting natural language prompts via personal devices. These prompts generate agent behaviors in a shared simulation space, visualized on a large screen. The images depict multiple stages of interaction, including prompt input (bottom left, bottom right), observation and discussion of emergent lifeform dynamics (top), and real-time feedback between users and the evolving system. The installation demonstrates how language-guided control supports collective exploration and co-creation of artificial life environments.
  }
  \label{fig:onsite_activity}
\end{figure*}

\section{Application: A Game of Ecosystem Construction}

To demonstrate the feasibility of our semantic complex systems framework, we developed an interactive game in which players collaboratively construct and evolve an artificial ecosystem. The experience unfolds in three stages, each reflecting a shift in system dynamics.

\textbf{Stage 1: Individual Symbolic Construction.}
Players access the game via a web interface and submit a natural language prompt, which is encoded by the Prompt2Param module into an initial Swarm configuration. This is sent via WebSocket to a Python-based evolution engine running CMA-ES, which returns real-time visualizations to the client. Players can pause evolution, revise prompts, and spawn new branches, iteratively shaping digital lifeforms that reflect their conceptual intent. This stage links symbolic language to agent-based dynamics, allowing abstract ideas to manifest as generative behavior.

\textbf{Stage 2: Multi-Agent Ecological Interaction.}
Once a lifeform is finalized, its behavioral parameters are transferred to a shared Unity-based simulation. Leveraging GPU compute shaders, the platform simulates tens of thousands of agents in real time. Lifeforms enter the virtual environment via network transmission and interact with others, merging, repelling, hybridizing, or asserting territorial boundaries. The simulation is broadcast live through OBS and Restream.io, supporting remote viewing and commentary. These behaviors emerge from both local agent rules and global stochastic effects. User-defined parameters not only shape individual agents but also influence the evolution of the shared environment, embedding conceptual intent into collective dynamics.

\textbf{Stage 3: Emergent Collective Regulation.}
As the ecosystem expands, the system records prompt histories, parameter trajectories, and behavioral outcomes. These are clustered and reduced via Principal Component Analysis (PCA) to extract semantic themes, conflict zones, and co-evolutionary patterns. Based on these insights, the system derives emergent “collective rules,” which represent higher-order behavioral logics shaped by dominant or negotiated tendencies among contributors. These meta-rules are re-injected into the simulation, affecting global dynamics beyond any single agent’s influence. In this final stage, semantic input reorganizes not only individual behavior but the generative logic of the ecosystem itself.

\section{Case Studies and User Scenarios}
We conducted two complementary user studies to evaluate both the semantic performance and participatory potential of our system.

\textbf{Controlled Evaluation of Semantic Guidance.}
The first study assessed the system’s ability to (1) generate agent behaviors aligned with natural language prompts, (2) support directional control through iterative prompt refinement, and (3) provide greater accessibility and expressiveness compared to manual parameter tuning. Ten participants completed three tasks in a controlled lab setting. In Task A, prompt-driven outputs guided by semantic feedback outperformed manually tuned results across three test prompts, measured by CLIP similarity scores, normalized user ratings, and preference counts (Figure~\ref{fig:user_study}A,B). In Task B, participants refined their prompts over three rounds, resulting in progressively higher CLIP alignment and satisfaction ratings (Figure~\ref{fig:user_study}C), demonstrating the effectiveness of language as an iterative control modality. Task C compared prompt-based interaction with manual tuning across five user-defined criteria, with language input consistently rated higher in creativity, controllability, and ease of use (Figure~\ref{fig:user_study}D). These findings support the system’s capacity for interpretable evolution while enhancing user experience in exploratory design.

\textbf{Public Deployment in a Generative Ecosystem.}
To evaluate the system in an open-ended setting, we deployed it in a participatory generative ecosystem game exhibited at the School of Arts and Design, Tsinghua University. This large-scale deployment, described in Section 5, demonstrated the framework’s effectiveness as a real-time creative platform and enabled observations of emergent social dynamics. Participants contributed prompts ranging from poetic abstractions to personal metaphors, with agents cohabiting and interacting in a shared simulation. The evolving ecosystem reflected both individual semantics and collective influence. These outcomes highlight the framework’s potential to support language-driven co-creation in participatory art and artificial life environments.

\section{Discussion}
Our findings suggest that semantic feedback provides an intuitive and expressive mechanism for interacting with complex generative systems. Compared to manual parameter adjustment, natural language enables users to convey abstract goals and conceptual metaphors without relying on technical knowledge. In our controlled study, participants reported higher satisfaction and perceived control when using prompt-based interaction, and the iterative refinement process allowed semantic intent to evolve in tandem with system behavior.

The public deployment further illustrated the potential of our framework as a platform for participatory generative design. As users contributed prompts and observed their agents interacting in a shared simulation, emergent dynamics began to reflect collective influences rather than isolated inputs. The integration of aggregated prompts into higher-level behavioral patterns demonstrated the system’s ability to synthesize distributed intent into evolving rule structures, supporting recursive, open-ended co-creation.

Building on this foundation, we plan to explore deeper language modeling for capturing metaphorical or ambiguous intent, expand the simulation framework to support richer ecological and aesthetic behaviors, and incorporate real-time user feedback directly into the optimization process. These directions will enable more nuanced semantic modulation, multi-agent co-evolution, and sustained ecosystem-level adaptation. We also envision applications in collaborative ALife simulation, generative worldbuilding, and educational contexts, where language operates not only as an interface but as an active medium for shaping generative systems.

\section*{Acknowledgments}
We gratefully acknowledge the Complex System Media Lab at Artineer LLC (USA) for research support in complex systems, and the School of Arts and Design at Tsinghua University for providing the exhibition venue for our participatory ecosystem game.

\begin{acks}
Thanks to Haoran Li from The Chinese University of Hong Kong for the technical support on Unity.
\end{acks}

\bibliographystyle{ACM-Reference-Format}
\bibliography{Reference}

\end{document}